\documentclass{sp}

\pdfoutput=1 

\pdfauthor{Scontras et al.}
\pdftitle{Practical introduction to RSA}
\pdfkeywords{Rational Speech Act models, probabilistic pragmatics, Gricean reasoning}

\title[Practical introduction to RSA]{A practical introduction to the Rational Speech Act modeling framework}

\author[]{
	\spauthor{Gregory Scontras \\ \institute{University of California, Irvine}} \AND
	\spauthor{Michael Henry Tessler \\ \institute{Massachusetts Institute of Technology}} \AND
	\spauthor{Michael Franke \\ \institute{University of Osnabr\"{u}ck}}
}

\usepackage{linguex}
\usepackage{qtree}
\qtreecenterfalse
\usepackage{tree-dvips}
\usepackage{phonetic}
\usepackage{xcolor}
\usepackage{pifont}
\usepackage{lineno}
\usepackage{graphicx}
\usepackage{booktabs}
\usepackage{multirow}
\usepackage{CJK}
\usepackage{tikz}
\usepackage{amsfonts}
\usepackage{amsmath}
\usepackage{hyperref}

\usepackage{listings}
\usepackage{color}
\definecolor{lightgray}{rgb}{.9,.9,.9}
\definecolor{darkgray}{rgb}{.4,.4,.4}
\definecolor{purple}{rgb}{0.65, 0.12, 0.82}
\definecolor{nc}{rgb}{0, 0.25, 0.5}

\usepackage{pxfonts}

\lstdefinelanguage{webppl}{
  keywords={typeof, new, true, false, catch, function, return, null, catch, switch, var, if, in, while, do, else, case, break, condition, factor, observe, Infer, adjust_score},
  keywordstyle=\color{black}\bfseries\textbf,
  ndkeywords={Bates_sample_function, speaker},
  ndkeywordstyle=\color{nc}\bfseries,
  identifierstyle=\color{black},
  sensitive=false,
  comment=[l]{//},
  morecomment=[s]{/*}{*/},
  commentstyle=\color{darkgray}\ttfamily,
  stringstyle=\color{red}\ttfamily,
  morestring=[b]',
  morestring=[b]"
}

\def\url#1{\expandafter\string\csname #1\endcsname}

\newcommand{\lam}{\ensuremath{\lambda}}
\newcommand{\sem}[1]{\ensuremath{[\![#1]\!]}}

\begin{document}

\maketitle

\begin{abstract}
	Recent advances in computational cognitive science (i.e., simulation-based probabilistic programs) have paved the way for significant progress in formal, implementable models of pragmatics. Rather than describing a pragmatic reasoning process in prose, these models formalize and implement one, deriving both qualitative and quantitative predictions of human behavior---predictions that consistently prove correct, demonstrating the viability and value of the framework. The current paper provides a practical introduction to and critical assessment of the Bayesian Rational Speech Act modeling framework, unpacking theoretical foundations, exploring technological innovations, and drawing connections to issues beyond current applications. 
\end{abstract}


\section{Introduction}

Much work in formal, compositional semantics follows the tradition of positing systematic but inflexible theories of meaning. In practice, however, the meanings listeners derive from language are heavily dependent on nearly all aspects of context, both linguistic and situational. To formally explain these nuanced aspects of meaning and better understand the compositional mechanism that delivers them, recent work in formal pragmatics recognizes semantics not as one of the final steps in meaning calculation, but rather as one of the first. Within the Bayesian Rational Speech Act (RSA) framework \citep{goodmanfrank2016,frankejaeger2016}, speakers and listeners reason about each other's reasoning about the literal interpretation of utterances. The resulting interpretation necessarily depends on the literal interpretation of an utterance, but is not necessarily wholly determined by it. This move---reasoning about likely interpretations---provides ready explanations for complex phenomena ranging from metaphor \citep{kaoetal2014metaphor} and hyperbole \citep{kaoetal2014} to the specification of thresholds in degree semantics \citep{lassitergoodman2013}.

The probabilistic pragmatics approach leverages the tools of structured probabilistic models formalized in a stochastic $\lambda$-calculus to develop and refine a general theory of communication. The framework synthesizes the knowledge and approaches from diverse areas---formal semantics, Bayesian models of reasoning under uncertainty, formal theories of measurement, philosophy of language, etc.---into an articulated theory of language in practice. These new tools yield improved empirical coverage and richer explanations for linguistic phenomena through the recognition of language as a means of communication, not merely a vacuum-sealed formal system. By subjecting the heretofore off-limits land of pragmatics to articulated formal models, the rapidly-growing body of research both informs pragmatic phenomena and enriches theories of linguistic meaning.

These models are particularly well-suited for capturing complex patterns of pragmatic reasoning, especially cases that integrate multiple sources of uncertainty. For example, in order to infer what a speaker has meant by an utterance in a given context, listeners may have to reason about lexical or syntactic ambiguity \citep{bergenetal2016,savinellietal2017,FrankeBergen2020:Theory-driven-s}, resolve underspecification in semantic meaning \citep{lassitergoodman2013}, or consider the possibility of non-literal interpretation \citep{kaoetal2014metaphor,kaoetal2014}. Listeners may also have to reason about the speaker's own likely epistemic state \citep{goodmanstuhlmuller2013,scontrasgoodman2017,HerbstrittFranke2019:Complex-probabi} or motives the speaker might entertain other than pure information exchange \citep{yoonetal2016,yoonetal2017}. Probability calculus provides a well-understood method for formalizing such complex patterns of reasoning, grounded in norms of rational belief formation and action choice. Recent advances in computer science provide accessible computational methods for implementing complex models of probabilistic reasoning, enabling us to naturally formulate and conveniently test pragmatic theories at a level of complexity and formal rigor that far exceeds what was possible before.

The current paper offers a practical introduction to the modeling framework, serving as a companion piece to the hands-on web-book at \href{https://www.problang.org}{www.problang.org} \citep{problang}. We begin in Section \ref{overview} with a high-level overview of RSA, walking through its basic implementation and the philosophical foundations that informed the architectural choices. We then explore variations to the basic architecture in Section \ref{variations}, surveying technological innovations that have allowed for broader empirical coverage. After a brief excursus on the relative benefits of ``thinking in math'' vs.~``thinking in code'' in Section \ref{prob-programs}, in Section \ref{practicalities} we discuss common practical considerations facing the modeler. In Section \ref{limitations}, we explore limitations of the current framework, which also serve as guidance for future extensions. Section \ref{summary} concludes.

\section{High-level overview of RSA} \label{overview}

The RSA framework views language understanding as a process of recursive social reasoning between speakers and listeners: listeners interpret the utterances they hear by reasoning about how speakers generate them; speakers choose their utterances by reasoning about how listeners interpret them. In the basic RSA model from \cite{frankgoodman2012} (hereafter, ``Vanilla RSA''), this recursion involves three layers of inference.\footnote{For an earlier proposal of this architecture, see \cite{BenzvanRooijOptimalAssertions2007}.} Typically formulated as statements of conditional probability, as in (\ref{L0}), (\ref{S1}), and (\ref{L1}), these inference layers correspond to models of speakers and listeners. We will go over these definitions step by step.

The reasoning grounds out in the naive, literal listener, $L_0$, who interprets utterances according to their literal semantics.
\begin{equation} \label{L0}
P_{L_0}(s \mid u) \propto P_{L_0}(s) \cdot \sem{u}(s)
\end{equation}
The semantics of utterance $u$ is captured in the meaning function $\sem{u} \colon s \mapsto [0;1]$, which maps states to truth values.
We assume binary truth-values here, but the formalism works just as well for non-binary, fuzzy values (e.g., \citealp{degenetal2020}).
Given a binary truth-conditional semantics, the rule in Equation~\eqref{L0} is equivalent to the following formulation in terms of belief update with the set of states where $u$ is true:
\begin{align}
  \label{eq:1}
  P_{L_{0}}(s \mid u) = P_{L_{0}}(s \mid \{ s \mid \sem{u}(s) = 1\})
\end{align}
According to Equation~\eqref{L0}, $L_0$ hears some utterance $u$, and updates their prior beliefs $P_{L_{0}}(s)$ about the world states $s$ with the information that $u$ is true.
If, as we will assume throughout unless noted otherwise, the prior beliefs of the literal listener are a uniform distribution (i.e., equal probability for all states), \eqref{L0} returns a uniform probability distribution over the states $s$ that $u$ maps to \texttt{true}.\footnote{For more discussion on using flat priors for the literal listener, at least in referential-communication games, see \cite{qingfranke2015}. Some models may wind up using an informative prior at the level of $L_0$, as in the context-inference models presented in Section \ref{context-inference} below.}

One layer up, a pragmatic speaker, $S_1$, chooses utterances in proportion to their utility $U_{S_{1}}$:
\begin{align} \label{S1}
  P_{S_1}(u \mid s) & \propto \exp (\alpha \cdot U_{S_1}(u;s))\text{, where} \\
  U_{S_1}(u; s) & = \textrm{log}P_{L_0}(s \mid u) - C(u) \nonumber
\end{align}
Utterances are useful to the extent that they maximize the probability that $L_0$ will infer the correct $s$ on the basis of $u$ (i.e., informativity), while minimizing the cost $C(u)$ of $u$ (speakers aim to be efficient). So, when selecting utterances, $S_1$ considers their effect on interpretation (i.e., on $L_0$'s resulting beliefs); utterances that are most likely to lead $L_0$ to the correct belief are most likely to be chosen by $S_1$.
The numerical utilities are mapped onto discrete choice probabilities by a soft-max function with parameter $\alpha$.
The larger $\alpha$ is, the more the speaker's choice probabilities converge to a strict maximization of utility.

At the top layer of inference, the pragmatic listener, $L_1$, interprets utterances to update their prior beliefs $P_{{L_1}}(s)$ by taking into account how likely the speaker would have been to produce the observed utterance $u$ in various states:
\begin{equation} \label{L1}
P_{L_1}(s \mid u) \propto P_{L_1}(s) \cdot \ P_{S_1}(u \mid s)
\end{equation}
In other words, unlike $L_0$, who reasons directly about the utterance semantics, $L_1$ reasons instead about the process that generated the utterance; that process is the speaker $S_1$. $L_1$ thus infers $s$ on the basis of $u$ by considering which states are a priori likely and reasoning about the probability that $S_1$ would have chosen $u$ to signal $s$ to $L_0$. Because $L_1$ reasons about $S_1$, who in turn reasons about the literal semantics in $L_0$, $L_1$'s interpretation is affected by the semantics of $u$, albeit only indirectly via the $S_1$ layer.

\begin{figure}[t]
\centering
\includegraphics[width=\textwidth]{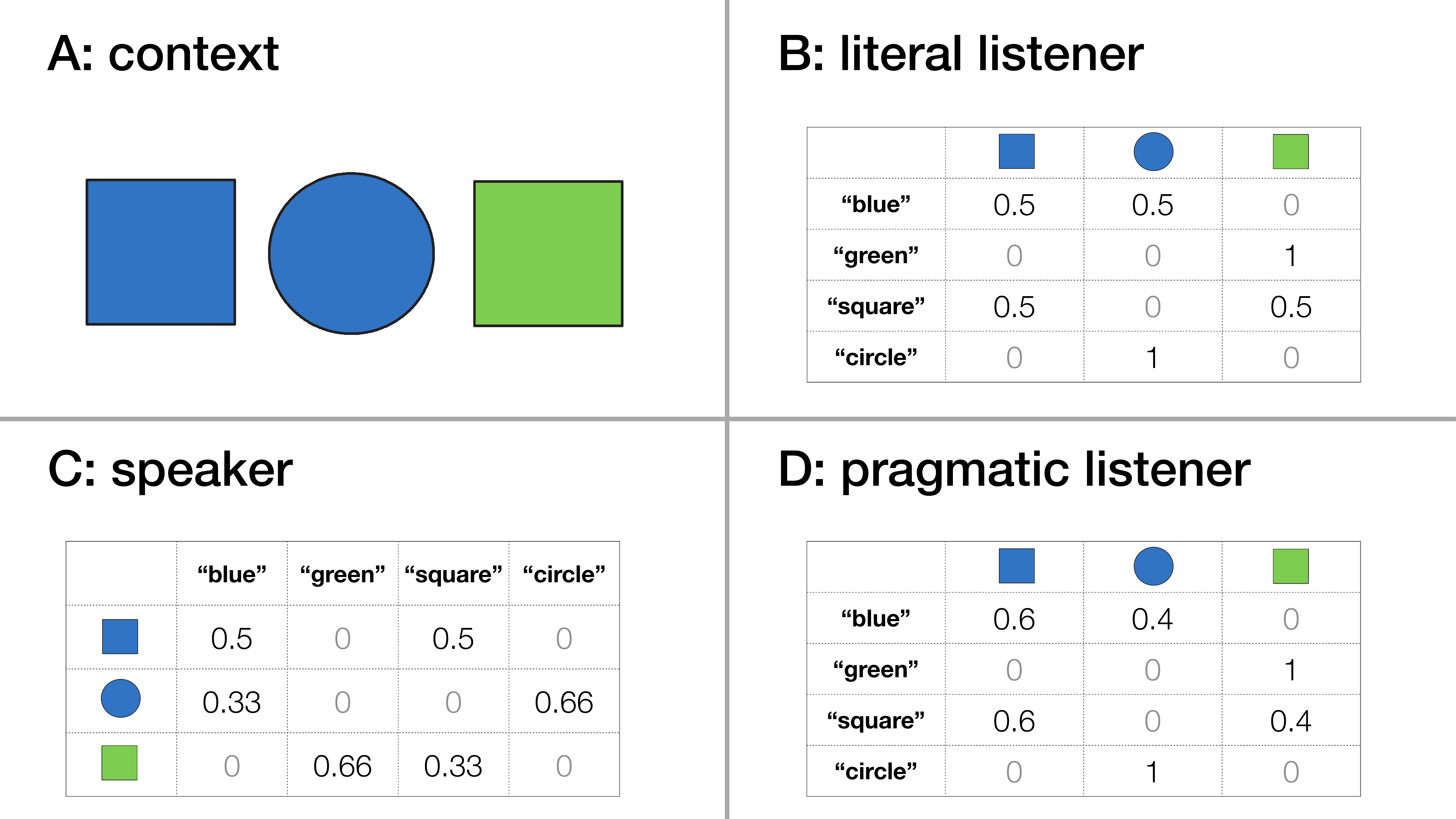}
\caption{An example of the Vanilla RSA model applied to a referential-communication game from \cite{frankgoodman2012}. Panel A shows the context of conversation with three potential referents. Panel B shows the interpretation probabilities of the literal listener. Panel C shows the production probabilities predicted for the pragmatic speaker when assuming $\alpha = 1$. Panel D shows the interpretation probabilities of the pragmatic listener, based on the production probabilities in panel C. All calculation assume flat priors, for the literal and the pragmatic listener.}
\label{ref-game}
\end{figure}

To understand the Vanilla RSA model better, it will help to consider a concrete example.
In its initial formulation, \cite{frankgoodman2012} used the RSA framework to model referent choices in efficient communication. Suppose we are in a context as in Figure \ref{ref-game} (panel A) with three objects: a blue square, a blue circle, and a green square. Suppose further that a speaker is trying to signal a single object in this world to a cooperative listener, and that the speaker can only use a single-word utterance to do so. The utterances available to the speaker are ``blue'', ``green'', ``square'', and ``circle''; the possible states the listener might infer correspond to the three objects: blue-square, blue-circle, and green-square. We have the expected truth-functional semantics for the utterances: ``blue'' maps blue-square and blue-circle to true but green-square to false, ``green'' maps blue-square and blue-circle to false but green-square to true, etc.

With the semantics as stated, $L_0$ interprets utterances to return uniform probability distributions over the compatible states (see Figure~\ref{ref-game} panel B). Hearing the utterance ``blue'', $L_0$ returns a belief distribution over states that divides probability equally between blue-square and blue-circle, the only objects compatible with the semantics of ``blue''. Hearing ``circle'', $L_0$ returns a distribution with 100\% of the probability on blue-circle---$L_0$ is certain that the speaker intends to signal blue-circle. With $L_0$ formulated in this way, we have an agent who interprets utterances naively, according to the literal semantics.

$S_1$ chooses utterances by simulating their effect on $L_0$. Suppose the speaker wants to communicate blue-circle to $L_0$. Two utterances, ``green'' and ``square'', stand no chance of communicating the intended state to $L_0$ and so they are ruled out entirely. The other two utterances, ``blue'' and ``circle'', are both literally compatible with blue-circle, but one of the utterances is more likely to lead $L_0$ to the correct belief state. If the speaker were to utter ``blue'', we saw that $L_0$'s belief distribution would be evenly split between blue-square and blue-circle. In other words, ``blue'' has a 50\% chance of leading $L_0$ to the correct state. On the other hand, ``circle'' has a 100\% chance of leading $L_0$ to the correct state. Assuming equal utterance costs, ``circle'' is thus twice as useful to the speaker as ``blue'' for communicating blue-circle to $L_0$, and $S_1$ reflects this asymmetry in utility by, if we assume $\alpha=1$, assigning twice as much probability to ``circle'' in the posterior distribution over utterance choices (see Figure~\ref{ref-game} panel C). For communicating the blue-square state, both ``blue'' and ``square'' have a 50\% chance of leading $L_0$ to the correct state, so both utterances are equally useful and thus equally probable in the $S_1$ posterior.

Within this simple reference-game scenario, $L_1$ reasons pragmatically about $S_1$ to break the symmetry in the semantics (see Figure~\ref{ref-game} panel D). Hearing ``blue'', $L_1$ will invert the $S_1$ model to determine which state (i.e., which object) the speaker is most likely trying to communicate. Had the speaker wanted to communicate the blue-circle state, we saw above that the speaker would have been more likely to utter ``circle''. But the speaker did not utter ``circle''; she uttered ``blue'' instead. This counterfactual reasoning leads $L_1$ to down-weight the probability of the state that ``circle'' would have uniquely picked out, since the speaker could have said ``circle'' but chose not to. As a result, more probability gets assigned to the blue-square state, the other state compatible with the semantics of the utterance. In this way, we capture the Gricean specificity implicature associated with uttering ``blue'' in a scenario as in Figure \ref{ref-game}: the speaker probably intends the blue square because if she wanted to communicate the blue circle she could have said ``circle''.

\paragraph{Prior beliefs \& pragmatic content}
A key component of the RSA framework is the updating of prior beliefs: listeners use the utterances they hear as signal with which to update their beliefs about the world.
In walking through the reference-game example above, we assumed a uniform prior over world states: before hearing the speaker's utterance, $L_1$ had no reason to suspect that any object was more or less likely to get referenced.
With a different, more informative prior, it is possible to shift the qualitative predictions of $L_1$.
Suppose that $L_1$ has reason to suspect that the blue-circle state is most likely to get referenced.
Now there are two opposing pressures operating on $L_1$'s interpretation of the utterance ``blue'' (i.e., on $L_1$'s posterior beliefs): the prior belief favoring blue-circle over blue-square, and the specificity inference discussed above that favors blue-square over blue-circle (if the speaker had wanted to reference blue-circle, she could have said ``circle'').
With a sufficiently strong prior in favor of blue-circle, it is possible that the most likely interpretation of ``blue'', according to an \emph{a priori} biased $L_1$, is blue-circle after all.

To dissect the impact of prior beliefs and genuine pragmatic interpretation at the level of the pragmatic listener, we must therefore compare the pragmatic listener's beliefs under a literal interpretation $P_{L_{1}}(\cdot \mid \{s
\mid \sem {u}(s)=1\})$ against the pragmatic listener's beliefs under a pragmatic interpretation $P_{L_{1}}(\cdot \mid u)$, both of which are probability distributions over states.
Following \cite{Skyrms2010:Signals}, we might look at the information about each state $s$ carried by the pragmatic interpretation of $u$ on top of its semantic interpretation, like so:
\begin{align*}
  \text{Info}(s, u) = { P_{L_{1}}(s \mid u) }
  -
  { P_{L_{1}}(s \mid \{s \mid \sem{u}(s)=1\}) }
\end{align*}
Dropping quantitative aspects from the picture, we may then say that \emph{utterance $u$ pragmatically conveys $s$} iff $\text{Info}(s,u) > 0$; that the \emph{pragmatic content conveyed by utterance $u$} is the set $\{s \mid \text{Info}(s,u) > 0\}$; and that an uttering of $u$ implicates the negation of $\{s \mid \text{Info}(s,u) < 0\}$.
The purpose of sketching these qualitative definitions is merely to underline how the RSA framework makes it possible that a listener computes a pragmatic inference but still puts strong credence in a state that is implicated to be false.

\paragraph{Relation to Gricean maxims.} The vanilla RSA model, as described above in Equations
\eqref{L0}--\eqref{L1}, is a direct formalization of Grice's idea that listeners can infer
pragmatic meaning based on the assumption that speakers adhere to certain rules of behavior,
the so-called ``Maxims of Conversation'' \citep{Grice1975:Logic-and-Conve}. This relationship to Grice's maxims becomes more
transparent if we rewrite the speaker's utterance-choice probability, starting from the
definition in \eqref{S1}.
\begin{align} \label{eq:S1-rewrite}
  P_{S_1}(u\mid s) & \propto \exp \left [ \alpha \left (\log P_{L_0}(s \mid u) - C(u) \right)  \right ] & \text{\textcolor{gray}{[definition \eqref{S1}]}} \\
  & = \left (P_{L_0}(s \mid u) \  \frac{1}{\log C(u)} \right)^{\alpha} & \text{\textcolor{gray}{[rules exponential \& log]}} \nonumber \\
  & = \left ( \frac{\sem{u}(s)}{\mid  \{s \mid \sem{u}(s)=1\} \mid } \ \frac{1}{\log C(u)} \right )^{\alpha} & \text{\textcolor{gray}{[definition \eqref{L0} \& uniform priors]}} \nonumber
\end{align}
The reformulation in~\eqref{eq:S1-rewrite} shows how the speaker's choice probabilities are a
product of three factors, each corresponding to a central postulate concerning how cooperative and, arguably, rational speakers should tailor their contributions
to a conversation. Suppose we recast the components as follows: (i) we realize that the expression $\sem{u}(s)$ just gives us the truth-value of $u$ in state $s$, so we write this explicitly as 
$\text{Truth}(u,s) = \sem{u}(s)$; we identify the expression $\mid \{s \mid \sem{u}(s)=1\} \mid ^{-1}$ as a measure of the logical strength of $u$, because it gives us the number of states in which $u$ is true, so that we explicitly write this quantity as  $\text{Informativity}(u) =\  \mid \{s \mid \sem{u}(s)=1\} \mid ^{-1}$; (iii) we consider the cost $C(u)$ of utterance $u$ as a measure of efficiency of economy to optimize during production, and so explicitly write $\text{Economy(u)} = \log C(u)$, using the logarithm for mathematical convenience. We can then rewrite the speaker rule as:\footnote{For
binary truth-values, $\alpha$ can be dropped from the factor $\text{Truth}$.}
\begin{align} \label{eq:S1-three-factor-formulation}
  P_{S_1(u\mid s)}   & \propto \text{Truth}(u,s) \ \text{Informativity}(u)^{\alpha} \ \text{Economy}(u)^{\alpha} 
\end{align}
These three factors directly capture (a formalization of) the Gricean maxims of Quality,
Quantity, and Manner. In effect, the RSA speaker assumes that speakers (i) maximize truth
(i.e., they utter no falsehoods if they can select at least one
true message; Quality), (ii) maximize informativity all else equal (Quantity), and (iii) maximize utterance
economy all else equal (Manner). The more rational a speaker (i.e., the higher the value of the scaling parameter $\alpha$), the more
pronounced the latter two optimization tendencies become. This behavior reflects the special
status that Grice bestowed on the Maxim of Quality: (binary) truthfulness is not affected by
the speaker's degree of optimizing informativity and costs.

\paragraph{Informativity from alignment of beliefs.} We saw above that the Vanilla RSA
model's definition of speakers' choice probabilities can be interpreted as a product of three
factors that correspond closely to Gricean maxims of Quality, Quantity, and Manner.
To better understand the motivation behind the speaker production rule in Equation~\eqref{S1}, we now show how it can be derived from an intuitive picture of belief alignment.
This formal exercise will also immediately pave the way for understanding variations of the RSA model (to be discussed below), which do not assume that the speaker knows the true world state \citep{goodmanstuhlmuller2013,scontrasgoodman2017,HerbstrittFranke2019:Complex-probabi}.

Suppose that the speaker has some belief about the relevant world states that gets captured in a probability distribution $P_{S_{1}\text{-}Bel}$;
after hearing utterance $u$, the literal listener's beliefs about the relevant world states are represented as another probability distribution $P_{L_{0}\text{-}Bel}(u)$.
An utterance $u$ is more useful than another utterance $u'$ to the extent that $P_{L_{0}\text{-}Bel}(u)$  (i.e., the listener's beliefs after hearing $u$) is more closely aligned with $P_{S_{1}\text{-}Bel}$ (i.e., the speaker's beliefs) than $P_{L_{0}\text{-}Bel}(u')$ (i.e., the listener's beliefs after hearing $u'$) is.
Vanilla RSA corresponds to an implementation of alignment in terms of an information-theoretic notion of divergence between probability distributions.
Behind it all is simply the idea that the speaker prefers utterances that best align speakers' and listeners' relevant beliefs about the world.

A useful notion of alignment of probability distributions is the information-theoretic
measure of Kullback-Leibler (KL) divergence. KL-divergence is not symmetric, but considers one of
two to-be-compared probability distributions as the ground-truth, or the objective function
to be approximated.
This asymmetry makes sense in language understanding because it is the speaker's
beliefs to which the listener should align. KL-divergence then measures divergence in terms of,
essentially, the expected number of extra bits needed to encode a signal that was generated
from the true distribution with the approximate distribution.
The Kullback-Leibler divergence between (baseline/true) probability distribution
$P_{S_{1}\text{-}Bel}$ and (approximate/to-be-optimized) probability distribution
$P_{L_{0}\text{-}Bel}(u)$ is defined as:
\begin{align}
  \label{eq:KL-divergence}
  \text{KL}(P_{S_{1}\text{-}Bel} \mid \mid P_{L_{0}\text{-}Bel}(u)) = - \sum_{s} P_{S_{1}\text{-}Bel}(s) \ \log \frac{P_{S_{1}\text{-}Bel}(s)}{P_{L_{0}\text{-}Bel}(s \mid u)}
\end{align}

Using KL-divergence, we can then state a more general definition of utterance utilities, to
replace the formulation in \eqref{S1}:
\begin{align}
  \label{eq:Utils-KL-based}
  U_{S_1}(u; s) = \text{KL}(P_{S_{1}\text{-}Bel} \mid \mid P_{L_{0}\text{-}Bel}(u)) - C(u)
\end{align}
The formulation in \eqref{S1} is a special case of the more general \eqref{eq:Utils-KL-based}
that follows if the speaker's beliefs about the relevant world states are degenerate, that is,
whenever the speaker knows the true world state $s^{*}$, so that $P_{S\text{-}Bel}(s^{*})=1$.
In that case, we have:
\begin{align*}
  \text{KL}(P_{S_{1}\text{-}Bel} \mid \mid P_{L_{0}\text{-}Bel}(u)) & = - \sum_{s} P_{S_{1}\text{-}Bel}(s) \ \log \frac{P_{S_{1}\text{-}Bel}(s)}{P_{L_{0}\text{-}Bel}(s \mid u)} \\
  & =  - \log\frac{1}{P_{L_{0}\text{-}Bel}(s^* \mid u)} = \log P_{L_{0}\text{-}Bel}(s^* \mid u)
\end{align*}
This derivation of the speaker production rule in Equation~\eqref{S1} in terms of belief alignment shows how to generalize the Vanilla RSA model to contexts in which the speaker might not be fully knowledgable about the true world state.

\section{Variations on Vanilla} \label{variations}

We have seen how Vanilla RSA can be used to model pragmatic reasoning in simple reference games, as well as specificity implicatures more generally. However, pragmatic reasoning involves much more complexity than Vanilla RSA captures. In particular, there are many different aspects of uncertainty that must be included in order to handle a broader range of language phenomena. In this section, we explore recent extensions of RSA meant to handle such complex language phenomena; each variation introduces reasoning about different sources of uncertainty:
\begin{itemize}
  \item contextually-relevant semantic parameters (e.g., degree thresholds; Section~\ref{meaning-inference}),
  \item the Question Under Discussion (Section~\ref{QUD-section}),
  \item aspects of the linguistic context (e.g., what counts as relevant or expected; Section~\ref{context-inference}),
  \item the speaker's access to knowledge (Section~\ref{sec:epistemic-inference}), and
  \item the speaker's utility calculus (e.g., whether it concerns informativity, politeness, etc.; Section~\ref{sec:compl-util-infer}).
\end{itemize}
We organize our discussion around deviations from Vanilla RSA, highlighting novel technology in language-understanding computation and the phenomena it captures. While we hope to cover a good deal of ground in this brief overview, it is important to note that the RSA modeling framework is an active area of research with new developments continually arising.

As a point of reference for Vanilla RSA, we repeat the $S_1$ utterance-choice rule below in (\ref{S1-repeat}), where a pragmatic speaker selects utterances in proportion to their utility in conveying some observed state of affairs to a naive listener who interprets the utterance according to its literal semantics---all while minimizing utterance cost. As we shall see, many of the extensions of RSA can be characterized in terms of their deviation from this simple utterance-choice rule.

\begin{align} \label{S1-repeat}
  P_{S_1}(u\mid s) & \propto \exp (\alpha (\log P_{L_0}(s \mid u) - C(u))) \text{, where} \\
  P_{L_0}(s \mid u) & \propto P_{L_O}(s)  \cdot \sem{u}(s) \nonumber
\end{align}

\subsection{Meaning inference} \label{meaning-inference}

Vanilla RSA assumes that the speaker and listener possess the same model of language, with a fixed, context-independent utterance semantics that is shared by both speakers and listeners. In this way, the speaker and listener are assumed to understand language in the same way regardless of the context in which the language is used. But what happens when there is uncertainty about the meanings of words, either because they are sensitive to context or because speakers and listeners are uncertain about the other's language model, such that the meaning of words might not be shared across the two? We can capture this flavor or uncertainty in RSA by parameterizing the interpretation function such that the utterance semantics is subject to change. This move is illustrated in \eqref{S1-lexical-uncertainty}, where the variable \textbf{x} modulates the truth-functional mapping for an utterance $u$. Once named, this interpretation-fixing variable \textbf{x} becomes subject to active pragmatic reasoning, resolved by the revised $L_1$ in (\ref{L1-lexical-uncertainty}). In other words, we can allow for uncertainty around $\textbf{x}$, such that the precise semantics of $u$ might be vague or underspecified in context. What results is the class of lexical uncertainty RSA models. 
\begin{align} \label{S1-lexical-uncertainty}
P_{S_1}(u\mid s, \textbf{x}) & \propto \exp (\alpha (\log P_{L_0}(s\mid u, \textbf{x}) - C(u))) \text{, where} \\
  P_{L_0}(s \mid u, \textbf{x}) & \propto P_{L_O}(s)  \cdot \sem{u}^{\textbf{x}}(s) \nonumber
\end{align}
\begin{equation} \label{L1-lexical-uncertainty}
P_{L_1}(s, \textbf{x} \mid u) \propto P_{S_1}(u \mid s, \textbf{x}) \cdot P(s) \cdot P(\textbf{x})
\end{equation}

\cite{lassitergoodman2013} use this technology to model the interpretation of vague gradable adjectives, whose meaning depends crucially on properties of the context, both linguistic and extra-linguistic. Take the adjective \emph{heavy}, which is true of some state (i.e., a weight) just in case the weight exceeds the relevant threshold for heaviness. However, we conclude that a \emph{heavy elephant} weighs substantially more than a \emph{heavy backpack}; moreover, the weight we attribute to the backpack is likely to vary depending on whether the speaker is a preschooler or an olympian. Thus, the threshold above which some state counts as heavy is unlikely to be a fixed, context-invariant value.

\citeauthor{lassitergoodman2013} handle the uncertainty around the threshold value by parameterizing the meaning of utterances containing gradable adjectives: \emph{heavy} will be true of some state just in case the weight exceeds the relevant threshold (as before), where this threshold (\textbf{x} in \eqref{S1-lexical-uncertainty}) gets fixed by pragmatic reasoning:
\begin{align}
\label{heavy-sem}
\sem{heavy}^{\textbf{x}} = \lam s. weight(s) > \textbf{x}
\end{align}
It is up to the pragmatic listener, $L_1$, to resolve the uncertainty surrounding \textbf{x}. To do so, $L_1$ hears the utterance and jointly infers both the state of the world (i.e., the relevant weight) and the relevant heaviness threshold. $L_1$ performs this inference just as in the vanilla model: which state and threshold would have been most likely to lead $S_1$ to select the utterance that $L_1$ encountered? In other words, $L_1$ simulates $S_1$'s behavior for the possible combinations of states and thresholds, then weights states and thresholds in proportion to the probability that they would lead $S_1$ to choose the utterance that was encountered.

A similar threshold-inference mechanism can be used to formalize the truth conditions of generic statements (e.g., \emph{Birds fly}), whose relation to the within-category property prevalence (i.e., the \% of birds that fly) has been historically hard to pin down (e.g., not all birds fly). 
Similar to the gradable adjectives model, the generics model adopts a threshold semantics where the generic is true just in case the prevalence of the feature $f$ within the category $k$ exceeds the relevant threshold:
\begin{align}
\label{generic-sem}
\sem{generic}^{\textbf{x}} = \lam f. \lam k. P(f \mid k) > \textbf{x}
\end{align}
Then, pragmatic reasoning can be used to jointly determine the prevalence  $P(f \mid k)$ as well as the threshold $\textbf{x}$ \citep{tesslergoodman2019}.\footnote{
\cite{tesslergoodman2019} actually use a reduced version of this model where the threshold inference occurs at the level of $L_0$, as opposed to $L_1$. \cite{goodmanlassiter2015handbook} argued via simulation that threshold inference at the level of $L_0$ results in interpretations that are too weak.
Their simulations, however, did not consider the kinds of state priors that are relevant for generics (see Section \ref{practicalities-prior} for further details on state priors for generics). 
It remains an open empirical question whether threshold inference at the level of the literal~vs.~pragmatic listener is preferred.
For related conceptual criticism of the approach of \cite{lassiter2017adjectival}, see also \cite{QingFranke2014:Gradable-Adject}.
}

This technology---parameterization of the utterance semantics subject to pragmatic reasoning---allows for a recasting of the division of labor between semantics and pragmatics. With lexical uncertainty, the semantic content of an utterance can be (at least partially) determined via pragmatic inference. \cite{bergenetal2016} seize on this innovation to model M-implicatures \citep{horn1984} and certain embedded implicatures \citep{hurford1974,chierchiaetal2012}. Rather than uncertainty around the semantics of a single utterance (e.g., a gradable adjective, as in the example above), \citeauthor{bergenetal2016}~assume uncertainty around entire lexica.
The lexical uncertainty approach to embedded implicatures is picked up, extended by, and evaluated alongside empirical data by \cite{PottsLassiter2016:Embedded-implic} and \cite{FrankeBergen2020:Theory-driven-s}.

Meaning inference also serves as a useful tool for modeling ambiguity resolution. \cite{scontrasgoodman2017} use an utterance-choice rule as in \eqref{S1-lexical-uncertainty} to model the resolution of distributive-collective ambiguities in plural predication. If a listener hears that \emph{the boxes are heavy}, the listener must jointly infer both the intended interpretation of the utterance (i.e., distributive, commenting on individual box weight, or collective, commenting on the total weight of the set) and the state of the world (i.e., the weights of the boxes). Returning to \eqref{S1-lexical-uncertainty}, \citeauthor{scontrasgoodman2017} treat \textbf{x} as the interpretation-resolving variable that determines whether $u$ receives a distributive or collective interpretation. \cite{savinellietal2017,savinellietal2018} use this technology to model the resolution of scope ambiguity, as in utterances like \emph{every horse didn't jump over the fence}. For \citeauthor{savinellietal2017}, \textbf{x} in \eqref{S1-lexical-uncertainty} serves to determine the relative scope of logical operators elements at LF. 
Both of these applications---plural predication and quantifier scope ambiguity---highlight how meaning inference in RSA can serve as a useful shorthand for providing computational-level descriptions of the process of ambiguity resolution. However, while these models offer a promising hypothesis for how listeners reason pragmatically to resolve ambiguity, it bears noting that neither application attempts to model the compositional processes that give rise to the relevant ambiguities in the first place. We return to this issue in Section \ref{limitations}.

\subsection{QUD inference} \label{QUD-section}

The technology we have considered up to now allows for the pragmatic enrichment of utterances: from ``blue'' to ``blue square'' or from vague ``expensive'' to ``expensive'' with a contextually-determined price threshold. However, given that $S_1$'s utility relies on successfully communicating the observed state to $L_0$ (i.e., on informativity), our enrichments can only ever be literal---we have no way for ``blue'' to be interpreted as ``green''.\footnote{
  Formally, this outcome is a direct consequence of grounding the speaker's production utilities in information theoretic surprisal and KL-divergence, as described in Section~\ref{overview}. The Vanilla RSA model, as based on rules \eqref{S1}, entails that if $\sem{u}(s)=0$, while the speaker's belief assigns positive probability to $s$, there is zero probability of the speaker choosing utterance $u$. In other words, Vanilla RSA predicts that the speaker never chooses an utterance when the speaker is not certain that the utterance is true of the relevant state.
}
But many instances of everyday language use are, strictly speaking, non-literal. For example, if you hear that someone paid ``a million dollars'' for their coffee at some hipster hangout, you are unlikely to infer that the actual price paid was \$1,000,000. Rather, you infer that the price was high and that the speaker is, to put it mildly, less than thrilled.

To allow for enrichments beyond the literal meaning, we must broaden $S_1$'s goals beyond informativity with respect to $s$. One way of conceiving of speaker goals is in terms of the Question Under Discussion (QUD) the speaker aims to answer. Under certain theories of pragmatics, all discourse proceeds with respect to some QUD \citep{roberts2012}; utterances in the discourse must at least partially answer the QUD in order to be cooperative and felicitous. By allowing for a broader range of QUDs and uncertainty around which QUD is intended, Vanilla RSA may be extended to capture uses of non-literal language.

As was the case with meaning inference in the previous subsection, we can illustrate this innovation by highlighting its effect on the vanilla utterance-choice rule. In \eqref{S1-QUD}, $S_1$'s utility continues to break down into informativity and economy components. However, now \textbf{x} determines what it is that $S_1$ endeavors to communicate to $L_0$. Viewing \textbf{x} as the QUD, $F(s,\textbf{x})$ serves to map $s$ to the answer to \textbf{x} that $S_1$ would like to communicate to $L_0$.
\begin{align} \label{S1-QUD}
P_{S_1}(u\mid s, \textbf{x}) & \propto \exp (\alpha (\log P_{L_0}(F(s,\textbf{x})\mid u) - C(u)))\text{, where} \\
  P_{L_0}(s \mid u) & \propto P_{L_O}(s)  \cdot \sem{u}(s) \text{ and } \nonumber \\
  F(s,\textbf{x}) \text{ is } & \text{a set of states} \nonumber 
\end{align}

\cite{kaoetal2014} use this technology to model hyperbole, as in the coffee-price example above. To see how, consider some possible states of the world about which a speaker might want to communicate:

\ex. \label{hyperbole-states}
\emph{Possible states of the world for the coffee-price scenario}:\\
$s_1$: \{\texttt{affect}: positive, \texttt{price}: \$3\}\\
$s_2$: \{\texttt{affect}: positive, \texttt{price}: \$7\}\\
$s_3$: \{\texttt{affect}: positive, \texttt{price}: \$1,000,000\}\\
$s_4$: \{\texttt{affect}: negative, \texttt{price}: \$3\}\\
$s_5$: \{\texttt{affect}: negative, \texttt{price}: \$7\}\\
$s_6$: \{\texttt{affect}: negative, \texttt{price}: \$1,000,000\}

Notice that the world states in \ref{hyperbole-states} have two properties: the \texttt{affect} of the speaker (i.e., whether the speaker feels positive or negative about the price paid), together with the actual \texttt{price} paid. A speaker talking about $s$ might have the goal of communicating about their \texttt{affect} and the \texttt{price} paid, or they may want to communicate only their \texttt{affect} or only the \texttt{price} paid. Viewed in terms of QUDs, these goals serve to partition the state space. With an \texttt{affect?} QUD, we partition the state space into two cells, corresponding to positive vs.~negative \texttt{affect}. With a \texttt{price?} QUD, we partition the state space into three cells, corresponding to the three possible \texttt{price}s. With a QUD asking after both \texttt{affect-and-price?}, we partition the state space into the six cells listed in \ref{hyperbole-states}.

With the utterance-choice rule in \eqref{S1-QUD}, $S_1$'s goals are QUD-specific: for communication to succeed, $L_0$ must simply arrive at the correct cell of the relevant partition; whatever $L_0$ infers about what happened within that partition is irrelevant to $S_1$'s goals. For example, if $S_1$ has a negative affect and is addressing the \texttt{affect?} QUD, utility depends on whether or not $L_0$ correctly arrives at the negative cell of the \texttt{affect?} partition, regardless of whether $L_0$ infers the full state to be $s_1$, $s_2$, or $s_3$. It is at this point where non-literal language suddenly becomes rational for a speaker. To see how, we have to say more about the prior knowledge speakers and listeners bring to bear on their communication scenarios.

World knowledge tells us that states in which the speaker actually paid \$1,000,000 (i.e., $s_3$ and $s_6$) are extraordinarily improbable, while states in which the coffee cost \$3 (i.e., $s_1$ and $s_4$) are the most likely; this knowledge gets reflected in $L_1$'s prior over world states, $P(s)$. We also have prior knowledge about how probable a speaker is to have a positive vs.~negative affect given a specific price paid for a cup of coffee: as prices increase, the probability of negative affect increases along with it. Now, return to a speaker addressing the \texttt{affect?} QUD: if $S_1$'s primary objective is to communicate negative affect to $L_0$, and, crucially, if the available utterances only concern price (i.e., ``I paid \$3/\$7/\$1,000,000''), then the rational thing for $S_1$ to say is that the coffee cost ``\$1,000,000''; given the extremely strong association between \$1,000,000 coffee and negative affect, $L_0$ is all but certain to arrive at the correct answer to the QUD: the speaker holds negative affect.

We have identified how non-literal utterances can be useful to a speaker, but we have yet to capture the non-literal aspect of their meaning. It is at the level of $L_1$ where utterances suddenly become non-literal. Hearing that the speaker paid ``\$1,000,000'', $L_1$ uses the informative priors on prices and their associated affects to arrive at a plausible interpretation. Given that coffee is sure to cost less than \$1,000,000, $L_1$ considers the possible QUDs the speaker might be addressing when selecting $u$. The \texttt{price?} and \texttt{price-and-affect?} QUDs are highly unlikely, given that \$1,000,000 is a highly unlikely price to pay for a cup of coffee. That leaves the \texttt{affect?} QUD. So, $L_1$ considers the possible full states that would have led $S_1$ to utter ``\$1,000,000'' in response to the \texttt{affect?} QUD. Via this counterfactual reasoning, $L_1$ arrives at a reasonable interpretation: $S_1$ is likely to have paid more than usual, so $s_1$ and $s_4$ are downweighted, but we are still on earth, so $s_3$ and $s_6$ remain highly unlikely. Of the remaining states, $S_1$ is most likely to have wanted to communicate negative affect with ``\$1,000,000'', so $L_1$ concludes that $s_5$ is more probable than $s_2$. In other words, $L_1$ hears that the speaker paid ``\$1,000,000'' for a cup of coffee and concludes (i) that the speaker is likely addressing the \texttt{affect?} QUD, (ii) that the speaker is likely trying to communicate negative affect, and (iii) that the price is higher than usual (which led to the negative affect) but still within the realm of possibility. This last conclusion---a hyperbolic interpretation of the price---is where non-literal meaning enters.

\cite{kaogoodman2015} use an extension of the hyperbole model to capture ironic interpretations of weather descriptions (e.g., ``It's terrible out'' to describe a beautiful sunny day). In addition to inferring speaker affect, the irony model incorporates an additional dimension of meaning that a speaker might want to communicate: their arousal about the weather (i.e., whether they feel strongly about it). The non-literal interpretation mechanism remains the same: if a speaker describes the weather as ``terrible'' but we know that terrible weather is extremely unlikely (the authors situate their imagined conversation in California, where the weather is rarely objectionable), then a listener will not conclude that the unlikely literal state (i.e., terrible weather) holds, but rather that the speaker must be addressing some other dimension of meaning, namely the strong arousal they feel in their positive affect toward the normal state of affairs: good weather. Similarly in \cite{kaoetal2014metaphor}, who model the interpretation of metaphors like ``John is a whale''. Rather than inferring that John is an aquatic mammal (a highly unlikely state of affairs), the listener concludes that the speaker chose the utterance to communicate about some other dimension of meaning, for example John's physical stature, which has whale-like properties.

\subsection{Context/Prior inference} \label{context-inference}

We now have technology to model vagueness and ambiguity on the basis of an underspecified semantics, as well as shifting goals and the non-literal language that addresses them on the basis of uncertainty around the QUD. So far, all of this reasoning has happened with respect to a fixed context, or common ground. What happens when a listener is unsure of the context a speaker has in mind in the production of their utterance? 
For example, upon hearing that a person is ``tall'', how does a listener identify the relevant comparison class against which to evaluate tallness? Does the speaker mean tall for a young child or tall for a basketball player?

In an analogous manner to the way we treat uncertainty regarding the literal meaning and uncertainty regarding the QUD, a listener can have uncertainty about (aspects of) the relevant context. In the case of ``tall'', perhaps the listener is unsure of the appropriate comparison class. Or maybe the listener is unsure of the facts of the world: what is the general prevalence of tall-ness? In (\ref{S1-context}) we amend the speaker choice function to reason about \textbf{x}, a variable that influences the relevant/likely states of the world $s$.

\begin{align} \label{S1-context}
P_{S_1}(u\mid s, \textbf{x}) & \propto \exp (\alpha (\log P_{L_0}(s \mid u, \textbf{x}) - C(u)))\text{, where} \\
  P_{L_0}(s \mid u, \textbf{x}) & \propto P_{L_0}(s \mid \textbf{x})  \cdot \sem{u}(s) \nonumber
\end{align}

Viewing \textbf{x} as the comparison class for a gradable adjective, we can build on the \citeauthor{lassitergoodman2013} adjectives model mentioned in Section \ref{meaning-inference} to describe the formal mechanism by which listeners reason about the context against which an adjective like \emph{tall} receives a context-specific interpretation. The adjectives model fixes the vague meaning of a gradable adjective by computing the relevant threshold via reference to the state prior, the prior distribution over heights within some comparison class. In a model that incorporates uncertainty about the comparison class, the prior over world states is treated as a conditional probability distribution that depends upon the comparison class \textbf{x}: $P_{L_0}(s\mid \textbf{x})$.

\cite{tessler2017comparisonclass} use this parameterization to model comparison class inferences when listeners hear only an adjective applied to an individual (e.g., when describing a basketball player, ``He is tall'').
The model assumes that the listener has access to the fact that the referent is a basketball player and knows that a basketball player is a person. Then, the listener's prior distribution of comparison classes is the set of all categories of which the referent is a member, weighted by the prior probability. The prior probability of different comparison classes is an unknown theoretical construct and the original comparison class inference model only uses two comparisons classes: the subordinate category (e.g., basketball players) and a basic or superordinate category (e.g., people). The inference proceeds by imagining the conditions under which a speaker would use the adjective heard (e.g., \emph{tall}) with each of the comparison classes (e.g., \emph{for a person}~vs.~\emph{for a basketball player}).

The uncertainty over the comparison class is a special case of uncertainty over contexts, about which an appropriately-constructed $L_1$ model can reason. \cite{degen2015wonky} use this technology to model the stubbornness of scalar inferences in the face of strong prior beliefs that would have made the implicature-calculated meaning false. In that study, the authors investigated listener interpretations of quantified statements involving ``some'' in a set of stimuli with diverse prior expectations concerning the probability of the \emph{all-state}. For example, ``John threw 15 marbles into the pool. Kevin, who observed what happened, said \emph{Some of the marbles sank.}'' In this scenario, the prior probability of a marble sinking in water is close to 1; thus, one would expect \emph{a priori} that all of the marbles would sink. However, the authors observe across multiple dependent measures that participants still draw the implicature---\emph{some but not all of the marbles sank}. \citeauthor{degen2015wonky}~account for this behavior using the context-inference mechanism that allows listeners to revise their prior beliefs when the utterances heard are sufficiently unexpected, positing a world that is somehow strange or \emph{wonky}---for example, a world in which marbles sometimes float.\footnote{
 If we construe the pragmatic content of an utterance of ``some'' as in Section~\ref{overview} in terms of the degree to which the utterance changes the pragmatic listener's beliefs about each state, it is still feasible to speak of a ``some but not all'' implicature of ``some'' even with heavily biased beliefs. In effect, this means that we should understand the \emph{wonky worlds} model not as an explanation of how the implicature persists in the light of heavy prior biases, but rather as addressing the mechanism by which, on top of the usual implicature, prior beliefs about the world can be revised.
}

\subsection{Epistemic inference}
\label{sec:epistemic-inference}

By now it should be clear that a productive way of extending the RSA framework is to take into account various sorts of uncertainty that enter into communication scenarios: uncertainty about the semantics, the goals, or the context itself. Another common source of uncertainty concerns the epistemic states of the agents who are communicating. Speakers commonly describe states of the world without total knowledge of those states. For example, a speaker might observe an empty plate on John's desk with pizza crumbs on it and describe the scenario with ``John ate some of the pizza.'' As far as the speaker knows, John could have eaten all of the pizza, but, without seeing the pizza box, the speaker lacks evidence to make this stronger claim. A hungry listener will interpret the speaker's utterance to infer just how much pizza John has eaten. Importantly, whether the speaker saw the pizza box or just the crumb-covered plate is likely to influence the listener's inference about the full state of the pizza. The mechanism involves reasoning about the knowledge that the speaker used to make her utterance (i.e., the speaker's epistemic state), and the implications that the knowledge has for likely states of the world.

We can model this inference by once again amending our speaker model.
Section~\ref{overview} already introduced a way of extending the speaker production rule in Equation~\eqref{S1} to also cover the case where the speaker is uncertain about the world state.
To allow the pragmatic listener to reason in a structured manner about the speaker's beliefs, we model the speaker's belief state $P_{S}(s \mid \textbf{x})$ as a function of some observation $\textbf{x}$ the speaker may have made to arrive at her beliefs.
The following formulation is equivalent to the previous formulation of speaker utilities in terms of KL-divergence, thus similarly motivated by the idea that the speaker favors utterances based on how closely they help align the beliefs of a literal listener with their own:
\begin{equation} \label{S1-epistemic}
P_{S_1}(u\mid \textbf{x}) \propto \exp (\alpha \mathbb{E}_{P(s\mid \textbf{x})}(\log P_{L_0}(s\mid u) - C(u)))
\end{equation}

\cite{goodmanstuhlmuller2013} use this approach to model rates of scalar implicature for the quantifier ``some.'' In the pizza example, the relevant inference concerns how likely it is for the listener to strengthen the ``some'' utterance to ``some but not all.'' \citeauthor{goodmanstuhlmuller2013} hypothesize that rates of scalar implicature should decrease as the speaker has less information that would verify whether the all-state obtains (e.g., whether John in fact ate all of the pizza); the authors present behavioral data supporting this claim. The amended RSA model offers an articulated analysis of why speaker knowledge access (and listeners' awareness of it) should have the effect it does on interpretation.

In the case of full knowledge access, $L_1$ reasons upon hearing ``some'' that if the speaker had wanted to communicate the all-state, she could have said ``all;'' but the speaker did not utter ``all,'' so she must know that the all-state does not hold. This reasoning leads $L_1$ to reassign the probability that would have been assigned to the all-state to the other states compatible with the semantics of ``some,'' namely the some-but-not-all states. This decrease in posterior probability assigned to the all-state serves as our index of scalar implicature. With partial knowledge access, the speaker may believe she is in the all-state, but she can never know for sure; similarly, upon observing only a plate with pizza crumbs, the speaker will never know that the all-state (i.e., that all the pizza was eaten) does \emph{not} hold. $L_1$ knows this about the speaker, and therefore is aware that, when hearing ``some'' from a speaker with partial knowledge access, the speaker is less likely to be in position to know that the all-state does not hold, thus canceling the counterfactual reasoning mechanism driving scalar implicature.

In their model of distributive-collective ambiguities in plural predication, \cite{scontrasgoodman2017} use this same sort of speaker knowledge manipulation to modulate rates of distributive interpretations. Hearing ``the boxes are heavy'' from a speaker who was unable to access the weights of individual boxes---because, for example, the speaker used a dolly to move all the boxes at once, and so accessed only their collective weight---a listener is less likely to interpret the utterance distributively, as commenting on the individual box weights. Rather, the listener is more likely to assign the utterance a collective interpretation, commenting on the total weight of the set of boxes. The reasoning comes straight from \cite{Grice1975:Logic-and-Conve}: ``Do not say that for which you lack adequate evidence.'' In \citeauthor{scontrasgoodman2017}'s model, the pragmatic listener is reasoning about a speaker who lacks the knowledge to verify a distributive interpretation on the basis of the (partial) observation she has made about the state of the boxes.

Once we have a model of variably uncertain speakers as in Equation~\eqref{S1-epistemic}, it is also feasible to investigate the pragmatic listener's inferences about the likely observation $\textbf{x}$ that helps explain the speaker's utterance.
Conceptually, this task amounts to inferring the speaker's level of expertise or competence.
While non-probabilistic models of Gricean inference often make reference to a monolithic (though defeasible) assumption of speaker competence \citep{Geurts2010:Quantity-Implic,vanRooijSchulz:ExhaustiveInterpretation,Spector2006:Scalar-Implicat}, holistic probabilistic inference allows a more gradient picture.
For example, after an utterance of a logically-stronger expression like ``all'', the pragmatic listener will put more credence in the proposition that the speaker has made an observation $\textbf{x}$ that makes the speaker knowledgable, than they would after hearing a logically-weaker expression like ``some''.

\subsection{Complex utility/utility inference}
\label{sec:compl-util-infer}
 
 So far, the different variations of RSA that we have considered all proceed with the assumption that speakers want to convey information to the listener; their utility concerns informativity with respect to the QUD.\footnote{Generalizations of the speaker's goals for communication exist. We could, for example, model a speaker's utility function in terms of closeness to the true state, for example, for state spaces with a clear similarity ordering \citep{Franke2014:Typical-use-of-}.} Though informational uses of language are arguably the bread-and-butter of a successful communicative system, speakers do not always say what they mean. We dawdle, prevaricate, and sometimes outright lie, and this behavior is often in the service of maintaining our social relationships. Perhaps rather than telling your friend that their haircut is ugly, you could say ``it's an interesting look''. Why do speakers deviate from choosing the maximally-informative utterance? We can model non-informational uses of language by augmenting the speaker's utility function: 
 \begin{equation} \label{S1-polite}
P_{S_1}(u \mid s, \textbf{x}) \propto \exp (\alpha  (
 \textbf{x} \cdot \log P_{L_0}(s \mid u) +
 (1 - \textbf{x}) \cdot  \mathbb{E}_{P_{L_0}(s \mid u)}[V(s)] - C(u)))
\end{equation}

Now, $S_1$ chooses an utterance on the basis of the true state $s$ (e.g., the speaker's true feelings about the haircut), where the standard RSA information utility is one component of the utility function. 
This informational utility is weighted by a mixture parameter $\textbf{x}$, where the second utility component receives the opposite weight ($1-\textbf{x}$).
The second component is what \cite{yoonetal2016} call \emph{social utility}, and is a function of the literal listener's subjective value $V$ of the state of the world they believe they are in given the utterance.
\cite{yoonetal2016} deploy this generalization of speaker utility to account for polite language use. 
It formalizes a version of \citeauthor{brown1987politeness}'s (\citeyear{brown1987politeness}) politeness theory, in which cooperative speakers have social goals to minimize any potential damage to the hearer’s (and the speaker’s own) self-image (called \emph{face}), in addition to standard informational goals.
Thus, $S_1$'s utility is a mixture of two-utilities: standard epistemic utility defined in (\ref{S1}) and a social utility: $U_{social}(u)  =  \mathbb{E}_{L_0(s \mid \sem{u})}[V(s)]$. 
This social utility term increases $S_1$'s production probabilities for utterances that convey states with high subjective value (e.g., ``it's \emph{beautiful}'', ``your talk was \emph{amazing}''), thereby incentivizing white lies. 

The pragmatic consequences of this complex utility function can be realized at the higher levels of pragmatic recursion.
The pragmatic listener $L_1$ can reason jointly about the state of the world and about the speaker's utility function (specifically, what is the speaker's trade-off between informativity vs.~social goals?). \cite{yoonetal2016} show that with information about the true state of the world and the utterance, the model can recover the speaker's likely goals, and, vice versa, with information about the speaker's goals and the utterance, the model recovers the likely true state of the world.

\section{Pragmatic agents as probabilistic programs} \label{prob-programs}

Our exposition of the framework so far has centered on descriptions of the models in terms of mathematical notation, more concretely in terms of formulae characterizing conditional probability functions (mass or density, depending on the case).
We tried to motivate and explain the ideas behind these formulae, for example by showing how the speaker's utility function in the Vanilla RSA model can be derived from a picture of communication in which the speaker is trying to minimize the distance (measured in information-theoretic terms) between the speaker's and a literal listener's beliefs.

While it is most common to communicate a probabilistic model in terms of such formulae-based representations in scientific writing, this approach is not the only way of conceptualizing a probabilistic model (be it an agent model, a statistical model, or otherwise).
Indeed, one very intuitive way of thinking about a probability distribution focuses not on a high-level, abstract description of the overall probability of events, but on the \emph{stochastic generative process} that we conceive of as the source of these events.
The web-book ``Probabilistic Language Understanding'', for which this paper is an introductory and complementary companion, uses probabilistic programs to implement RSA-style probabilistic agent behavior in terms of explicitly-formalized stochastic generative processes \citep{problang}.
The main motivation of this section is to: (i) highlight the distinction between ``thinking in math'' and ``thinking in code'' when it comes to developing probabilistic models, (ii) to demonstrate the usefulness of ``thinking in code'' and thereby underlining the added value of the web-book, and (iii) to argue that, ideally, researchers should develop models by flexibly exploring conceptually-reasonable ideas in terms of both a mathematical and a code-based algorithmic perspective.

To see the difference between a mathematical and a process-based representation of a probability distribution, consider the following definition of a probability density function on some $x \in [a;b]$ parameterized by some integer-valued $n>0$:
\begin{align*}
  P(x \ ; \ n) =
  \begin{cases}
    \sum_{k=0}^{n}(-1)^{k}\binom{k}{n} \left( \frac{x-a}{b-a} - \frac{k}{n} \right)^{n-1} \textrm{sgn} \left( \frac{x-a}{b-a} - \frac{k}{n} \right) & \text{if } x \in [a;b]\\
    0 & \text{otherwise}.
  \end{cases}
\end{align*}
Unless you are already intimately familiar with the Bates distribution, chances are that this mathematical representation tells you very little about what is captured here.
Yet, despite the complex math, the Bates distribution derives from a very simple procedural idea:
sample $n$ times from a uniform distribution in the interval $[a;b]$ and look at the mean $x$ over all of these $n$ values.
The Bates distribution gives the probability of sampling any concrete value of $x$ by that process.
The upshot of this example is that a formula-based mathematical representation of a probability distribution can be highly useful for some applications, but it might not be useful for all purposes, one such purpose being to think about a possible process that could generate a particular output in a certain (stochastic) manner.

Probabilistic programming languages like WebPPL \citep{goodmanstuhlmuller2014}, in which the examples from the accompanying web-book are written, help to make the process-based perspective explicit.
Complex probability distributions, even those for which no closed-form mathematical representation is known, can be approximately represented by a large-enough set of representative samples.
These approximate representations are good enough for most common practical purposes.
Even more useful than a large set of fixed samples is a function that returns samples every time it is called.
Considerations like these are why probabilistic programming languages like WebPPL allow the user to define probability distributions in terms of a \emph{sampling function}, formulated simply as a process that would generate a single sample (if executed once).
For example, the Bates distribution could be written down in WebPPL-inspired pseudo-code as follows: 
\begin{lstlisting}
var Bates_sampling_function = function(n, a, b) {
  // take n samples from a uniform distribution on [a;b]
  var unif_samples = sample_from_uniform(n, a, b)
  return(mean(unif_samples))
}
\end{lstlisting}
This sampling function is sufficient to work with the Bates distribution (at least for most common applications).
Indeed, the web-book defines the RSA speaker and listener protocols, which we explicated above in terms of formulae, in exactly these kinds of sampling functions.

Adopting a perspective on language production and interpretation in terms of sampling functions is what we here refer to as ``thinking in code.''
To ``think in code'' is particularly useful in the context of cognitive modeling because it enables a sampling-based perspective of an agent's reasoning and decision-making process in line with recent accounts of resource-bounded rationality \citep{GriffithsLieder2015:Rational-Use-of, lieder2020resource}.
We will walk through two example cases where ``thinking in code'' and the adoption of a resource-bounded rationality perspective can add interesting flexibility to the modeling of pragmatic agents.

\subsection{Utterance priors \& utterance salience}

The Vanilla RSA model defines the speaker's utterance-choice probability in terms of \emph{utterance costs}, $C(u)$, which we identified as a potential formalization of Gricean Manner or a general constraint of \emph{utterance economy}.
Here is the original definition and the suggested refactorization from Section~\ref{overview}:
\begin{align} \label{S1-repeat-again}
  P_{S_1}(u\mid s) \propto & \exp (\alpha (\log P_{L_0}(s \mid u) - C(u)))  \\
       & = \text{Truth}(u,s) \ \ \text{Informativity}(u)^{\alpha} \ \ \text{Economy}(u)^{\alpha} \nonumber
\end{align}
In this formulation, the cost $C(u)$ of an utterance $u$ affects the speaker's choice of expression: the higher the cost of $u$, all else equal, the lower the probability that the speaker picks this utterance.
This design is motivated by the idea that speakers economize production effort, preferring to utter shorter or otherwise more efficient words or phrases.
Consequently, in line with this interpretation of minimization of utterance effort, the strength of the impact of utterance costs is modulated by $\alpha$: the higher the value of $\alpha$, the stronger the speaker's tendency to minimize utterance costs, all else equal.
This outcome is intuitive if we think of $\alpha$ as a rationality parameter:
the more rational an agent, the more sensitive that agent should be about the minimization of effort.

Adopting a sampling-based perspective inspired by resource-bounded rationality, we can conceive of a natural alternative to the speaker model in \eqref{S1-repeat-again}, in which differences between utterances are not treated as costs to be rationally economized, but rather as differences in salience---that is, the ease with which utterances comes to mind.
The latter factor of \emph{differential salience} is arguably not subject to optimization: it is not the case that the more rational a speaker is, the more she would tend to select only the utterances that come to mind most easily.
We can implement differential salience (ease of retrieval) with a mathematical formulation as follows:
\begin{align}
  \label{eq:S1-utterance-salience}
  P_{S_1}^{\text{Salience}}(u\mid s)   & \propto \text{Truth}(u,s) \ \text{Informativity}(u)^{\alpha} \ \text{Salience}(u)
\end{align}
The speaker model in \eqref{eq:S1-utterance-salience} includes the factor $\text{Salience}(u)$ as an \emph{utterance prior}, suggesting an algorithmic picture of utterance choice: the speaker searches for utterances to choose, based on a gradient of how easily these utterances come to mind, then compares the available options (weighted by their salience/prominence) based on the other factors relevant for communication: truth and informativity (see \citealp{Tielvan-TielFranke2021:Probabilistic-p} for an example application of salience priors to the choice of quantity expressions).

The salience-based speaker model in \eqref{eq:S1-utterance-salience} is implemented very naturally in WebPPL (or other probabilistic programming languages); WebPPL-inspired pseudo-code looks as follows:
\begin{lstlisting}
var speaker = function(state) {
  // sample an utterance from an utterance prior
  //   (= 'see what comes to mind')
  var utt = sample_from_utterance_prior()
  // weigh sampled utterance by how true and informative it is
  adjust_score(truth(utt, state) * informativity(utt)^alpha)
  // return utterance
  return(utt)
}
\end{lstlisting}
Notice that, for technical reasons, WebPPL indeed \emph{requires} the specification of an utterance prior even if we do not use it to implement salience.\footnote{WebPPL requires the inclusion of an utterance prior because it minimally needs to know the set of all available utterances. If no \emph{differential salience} is to be modeled, the utterance prior in a WebPPL program must still be specified but should then just be uniform.}

\subsection{Uncertainty from sampling}

Another example of a case where ``thinking in code'' makes a plausible conceptual idea much more readily available than a formula-based approach arises when we model a speaker who is uncertain about the true state of the world.
In Section~\ref{sec:epistemic-inference}, we described the model of \cite{goodmanstuhlmuller2013}, the speaker component of which is repeated here:
\begin{equation} \label{S1-epistemic-repeat}
P_{S_1}(u\mid \textbf{x}) \propto \exp (\alpha \mathbb{E}_{P(s\mid \textbf{x})}(\log P_{L_0}(s\mid u) - C(u)))
\end{equation}
The speaker is assumed to choose utterances proportionally to how much they minimize the Kullback-Leibler divergence between the speaker's own belief and the literal listener's belief after each utterance.
This model is easily written out in concise mathematical notation like in \autoref{S1-epistemic-repeat}, but the WebPPL code to implement it is rather clumsy (see accompanying web-book) and its calculation can be very costly for large sets of possible world states.
However, ``thinking in code'' and conceptualizing a resource-limited reasoner suggests a different model, one that is difficult to write down in mathematical notation but that is much easier to write down in a probabilistic program---and also much more efficient to approximate.
This alternative speaker model looks as follows in WebPPL-like pseudo-code, where \verb|belief_state| is a probability distribution capturing the speaker's belief state:
\begin{lstlisting}
var speaker = function(belief_state) {
  // sample an utterance from an utterance prior
  var utt = sample_from_utterance_prior()
  // sample a possible world state from the speaker's beliefs
  var state = belief_state()
  // weigh sampled utterance by how true and informative it is
  //   based on the current sample of a possible world state
  adjust_score(truth(utt, state) * informativity(utt)^alpha)
  // return utterance
  return(utt)
}
\end{lstlisting}
This algorithmic speaker model imagines that, instead of globally reasoning about all possible world states and weighing each one in terms of how probable they are to the speaker, the speaker just samples a random world state from her beliefs and then chooses a true and informative utterance for the sampled state.\footnote{In fact, it is this easier model that \citet{goodmanstuhlmuller2013} implemented, as is apparent from the plots of the model's predictions given in the paper. The key difference between the KL-based and the sampling-based model is that the former will never have an agent choose an utterance for which she cannot rule out that it might be false. See the accompanying web-book for more on this topic.}

\bigskip

\noindent This example serves to highlight how the available tools---being able to both ``think in math'' and to ``think in code''---help to shape the resulting theories. By having a multiplicity of tools handy, the modeler enlarges the space of conceivable (and practically-realizable) theories. While our focus in the current paper is RSA from a mathematical perspective, much of the work in implementing an RSA model involves conceiving of the relevant agents as sampling-based probabilistic programs; the accompanying web-book offers a hands-on introduction to this skill. In the following section, we dive more deeply into the practical considerations faced in the design and implementation of an RSA model.

\section{Modeling practicalities} \label{practicalities}

Section \ref{variations} presented a menu of options to the would-be modeler: given a language-understanding phenomenon, which pieces of technology are likely to prove most useful in a formal treatment? Non-literal phenomena might benefit from a treatment via QUD inference or complex utility; vague or underspecified language suggests meaning inference, combined perhaps with uncertainty about the relevant context/prior; in cases where the speaker is unknowledgeable or behaves in an unexpected manner, we can try epistemic inference and/or context inference. Having introduced the basics of the RSA modeling framework, as well as useful additions and amendments, our focus now turns to the practical considerations involved in designing and implementing an RSA model, and testing the predictions of that model against human behavior.

We focus in this section on practical issues arising during model development and testing. Concretely, we will address questions like: which discrete choice options (e.g., world states and utterances) to include in a model; which priors to assign to world states; how to set other (numerical) model parameters (e.g., the rationality parameter $\alpha$); how to link model predictions and empirical data for (Bayesian) data analysis.

\subsection{Determining world states, utterances etc.} \label{practicalities-prior}

One of the most important modeling decisions concerns setting up the conversational scenario: how do we represent the states of the world and utterances that can be described in our model? In other words, what constitutes a world-state, and at what level of linguistic description will utterances be treated? Answering these questions often entails specifying the universe of possibilities of states and utterances (i.e., the support of the prior distributions). 
In the reference game from \cite{frankgoodman2012} depicted in Figure \ref{ref-game}, we equate states with individual objects, and the set of possible states corresponds to the set of objects present in the visual array. But only certain features of the objects are relevant: their shape and color. Properties like size or position do not enter into the reasoning about shape and color in this reference game (or at least we do not think they do), and so they need not enter into the representation of states in the model. In \ref{hyperbole-states}, we saw possible states from the hyperbole model of \cite{kaoetal2014}; of all the various facts of the world, the authors narrowed in on two for their model: the price paid for some object and the valence of the affect of the person who paid it. 
But the authors did not consider all possible prices ranging from \$0  -- \$5 trillion (roughly the total amount of money in the world); they narrowed the possible prices to a set that is probably reasonable for the conversational scenario at hand.

One of the more ingenious---and abstract---state implementations comes from the generics model \citep{tesslergoodman2019}. Generic statements update beliefs about the prevalence of a feature in a category (e.g., the prevalence of egg-laying among robins upon hearing that ``robins lay eggs''). The prior, then, is a distribution over possible prevalences. \citeauthor{tesslergoodman2019} treat this distribution as a distribution over alternative kinds (e.g., bluejays, horses, tarantulas, etc.), akin to the implementation of the comparison class in the vague adjectives models \citep{lassitergoodman2013, tessler2017comparisonclass}. Each kind is then modeled as having its prevalence for the relevant property.

The choice of what possibilities to include is relevant to every aspect of the model where alternatives must be specified: alternative states, alternative utterances, alternative QUDs, etc.
There are several options for determining these internal parameters.
First, if a model's foremost purpose is the analysis of data from a concrete experiment, then it may be that the stimuli and available choice options of the experiment suggest, if not necessitate, a precise set of states, utterances, or other model ingredients.
The experiments of \citet{frankgoodman2012}, for example, present exactly three objects as potential referents and offer only a small set of discrete choice options for utterances in each trial.
So here, and in many other cases, the logical design of the experiment to be modeled dictates the nature and scope of some or all internal parameters.

Without obvious constraints from an experimental setup (or in order to determine those constraints), another reasonable approach is to incorporate an experiment in the model-development pipeline to determine empirically what reasonable state distinctions might be. For example, in developing their model of metaphors like ``John is a whale'', \cite{kaoetal2014metaphor} began by eliciting the most salient features for a set of animals (e.g., \emph{large}, \emph{graceful}, and \emph{majestic} in the case of whales).
Another possibility is, when dealing with numerical world states (e.g., in the interpretation of adjectives or other vague expressions), to use a pre-test to elicit likely ranges of numbers for particular experimental items \citep[e.g.,][]{FrankeScholler2016:Semantic-values}.

Determining the set of alternative utterances is an open issue in pragmatic theories broadly \citep{ChemlaSingh2014:Remarks-on-the-,Katzir2007:Structurally-De,FoxKatzir2011:On-the-Characte}, and the literature on RSA has not yet arrived at a set of best practices for determining alternative utterances.
A few general considerations can still be mentioned.
First, where linguistic theory makes clear indications of which alternatives are reasonable (e.g., from lexical association or from focus mechanisms), the RSA modeler should take this insight into account.
In some domains of application, however, no such clear predictions exist (e.g., for situated referential language, generics, or tropes).
In these cases, it is important to minimally ensure that the presented RSA model is \emph{prima facie} plausible as a representation of the speaker's or listener's subjective conceptualization of the decision situation (production or interpretation, depending on what we focus on; \citealp{Franke2012:Pragmatic-Reaso}).
For example, this perspective on what an RSA model represents can make it plausible to also include an informationless ``silent'' or ``null'' utterance.
Conceptually, a ``null'' utterance can be plausible in cases where speakers may reasonably be assumed not to want to say anything at all.
Technically, a ``null'' utterance can be useful for understanding the meaning of an utterance even without reasoning about explicit alternatives (e.g., in the study of vagueness; \citealp{lassitergoodman2013}).

\subsubsection{Prior probabilities of world states}

Having decided what the relevant alternatives are and how to model them, the task then turns to implementing the appropriate priors: beliefs about which alternatives are more or less likely a priori. Some priors can be estimated empirically. With state priors, it is sometimes possible to ask people explicitly about the relevant probabilities. For example, \cite{kaoetal2014} asked participants to rate the probability of paying various prices for certain objects and how likely it is that someone would consider a given price expensive; those ratings were normalized and served as the state prior probabilities in the hyperbole model. In the reference game model, \cite{frankgoodman2012} got at the state prior probabilities using a less direct method. To determine which objects in Figure \ref{ref-game} were more likely to get referenced a priori, the authors told participants that someone had used an unknown word to signal one of the objects; based on participants' guesses at which object the speaker was intending, \citeauthor{frankgoodman2012} were able to estimate the relevant state prior.
In the experimental tests of the generics model, \cite{tesslergoodman2019} employ two alternative techniques: one has participants freely generate categories and then rate the prevalence of a target feature for those categories; another has participants answer targeted questions about the abstract parameters of the prior distribution and then uses Bayesian data analysis to reconstruct the appropriate priors. 

In some cases---for example, an analytic demonstration of a qualitative prediction---prior estimation might be unnecessary or infeasible. In such cases, one ought to make minimal assumptions, for example by assuming an uninformative, uniform prior. Better yet, the modeler may want to demonstrate the qualitative predictions under a number of different assumptions about the shape of the prior to show the model's (in)sensitivities to choice of prior.

\subsection{Further (numerical) parameters}

Like the specific probabilities assigned in the various priors, other free parameters of the model must be fixed in order to generate predictions. Free parameters common to RSA include $\alpha$, which controls speaker optimality, and the various utterance cost parameters. These parameters are often not of direct theoretical interest but sometimes can be informed by domain knowledge. For instance, it is often natural to assume that the cost of an utterance is a function of either its length  (which could be measured in terms of number of words, syllables, or characters) or that the salience of an utterance is related to its frequency (estimated from a corpus).

Unfortunately,  no comparable motivation exists for the setting of the speaker optimality parameter $\alpha$, and efforts to convince a participant that a speaker is more or less rational may lead to unintended pragmatic inferences. For these reasons, the value of $\alpha$ is generally not directly theoretically interpretable. In practice, however, $\alpha$ tends to be greater than or equal to 1, and not very high ($\lessapprox 50$). The interpretation of an $\alpha$ less than 0 is that (the listener believes that) the speaker behaves irrationally, preferably chosing the options which minimize expected utility. If such a value of $\alpha$ is needed to obtain otherwise reasonable model predictions, this is a signal that the model is in some way mis-specified, which may prompt a closer look at the state and utterance parameterization.  See \cite{zaslavsky2020rate} for a fuller discussion of the role of $\alpha$ in RSA.

For the Vanilla RSA model, the number of layers of recursion can also be thought of as a free parameter. Empirically, it has been observed that levels of recursion and speaker optimality trade-off with one another  and hence lead to unidentifiability of the parameters \citep{frank2016rational}. Recent analytic analyses have shown, however, that speaker optimality determines a tradeoff between the expected utility optimized by RSA recursion and a conception of communicative effort \citep{zaslavsky2020rate}.

\subsection{Linking functions}

A fully-specified RSA model makes qualitative, probabilistic predictions that can be tested against empirical data.
Interpreting those predictions, however, requires a clear linking function from the model output to human behavior, and the specific linking function needed will depend on the human behavior to be modeled.

RSA returns predictions in terms of the probability of either a state and/or other variables inferred by the listener (e.g., QUDs, thresholds, etc.) or a choice of expression or utterance (for speaker models).
Thus, if the behavioral data is in the form of a particular state or a particular utterance (e.g., in a forced-choice task), then the model can be related to the data directly; the model provides the probability of the observed outcomes and it is possible to compute statistics such as the likelihood of the data under the model (and compare to the likelihood of the data under different models). 

In cases where the response variable is not a particular state or utterance, a linking function will need to be specified. One common response variable for experimental semantics/pragmatics researchers is a truth-value judgment \citep{crainmckee1985,crainthornton1998}. These judgments are used to identify the situations that a sentence can truthfully describe. Assuming that the truth conditions of a sentence---what it takes to make the sentence true---are at least constitutive of sentence meaning \citep[e.g.,][]{chierchiamcconnellginet2000}, then truth-value judgments inform how it is that speakers understand sentences. But how can we model such judgments? Given that truth-value judgments are used to inform our understanding of sentences, there is a temptation to treat the process that goes into generating these judgments as one of language comprehension, corresponding to the listener layer of an RSA model \citep[e.g.,][]{PottsLassiter2016:Embedded-implic}. However, several authors have suggested that truth-value judgments are better modeled as a form of language production: whether a speaker would use the sentence to describe a given scenario (e.g., \citealp{degengoodman2014,Franke2016:Task-types-link,savinellietal2017,tesslergoodman2019,jasbietal2019}). In RSA terms, this decision links to the probability of uttering the sentence of interest given some observed state---predictions from a speaker layer of the model. In models where the $S_1$ predictions depend on more than merely the observed state (as in the extensions discussed Section~\ref{variations}), generating speaker predictions that correspond to truth-value judgments may require an additional layer of recursion: an $S_2$ who observes a state $s$ and chooses an utterance to communicate $s$ to $L_1$, who in turn does the work of resolving any additional variables that $S_1$'s calculation uses (e.g., the interpretation-resolving \textbf{x} in (\ref{S1-lexical-uncertainty}) and (\ref{L1-lexical-uncertainty}); \citealp{savinellietal2017}).

Other common experimental measures include slider-bar or likert-scale ratings (e.g., of the probability of a state) or betting procedures (e.g., eliciting the probability of all states simultaneously as in \citealp{frankgoodman2012}). 
Though there is the temptation to interpret these measures as directly relatable to the model output (e.g., they may both be treated as probabilities which sum to 1), the correct way of handling these data is by articulating a linking function.
For example, likert-scale ratings should ideally be analyzed with a cumulative logit or cumulative probit linking function used for ordinal regression \citep{Franke2014:Typical-use-of-,Franke2016:Task-types-link}.
Or, when a participant places ratings via sliders on all states, a logit-transformed Gaussian may be an appropriate link so that the model places a probability on the actual responses that participants provide (see \citealp{franke2016does} for an analysis of different linking functions in the context of a prior-elicitation task).

\subsection{Parameter estimation and model comparison}

Besides actually building the RSA model of a linguistic phenomenon, the modeler is often tasked with evaluating (or criticizing) the model, as well as comparing it to alternative models.
These evaluations and comparisons may be informal when examining whether relevant qualitative phenomena (e.g., symmetry-breaking) are exhibited by a model.
Comparisons and evaluations, however, like the modeling of the pragmatic phenomenon itself, can be formalized. 
In this regards, Bayesian data-analytic methods are regarded as the gold-standard, in addition to being theoretically-synergistic with the Bayesian pragmatic views of language understanding.
Instead of a listener asking ``what did a speaker likely intend given their utterance (and my internal model of the speaker)?'', the question of parameter estimation for the scientist is ``what are credible values of the model's parameters given the data we've observed (and our model of the data [i.e., the RSA model])?''
Many resources are available for learning Bayesian data analysis; we recommend: \cite{gelman2013bayesian, kruschke2014doing, lee2014bayesian,Lambert2018:A-Students-Guid}.

For each of the above modeling practicalities, the modeler will find herself with additional decisions about variables of the model. For example, the speaker optimality parameter $\alpha$ is generally considered a free parameter of the model that may be fit to the empirical data (i.e., it adds a degree-of-freedom to the model). The most straightforward Bayesian approach to addressing what value to set for $\alpha$ involves setting a prior over plausible values of that parameter and then performing Bayesian inference, conditioning on the observed behavioral data in order to infer what  the most believable values of $\alpha$ are given the RSA model and the observed data.
Plausible \emph{a priori} values of parameters can be gleaned by reviewing other related RSA models. For example, at the time of this writing, we observe that most RSA models use $\alpha$ values between 0 and 20, though the actual plausible range of values will differ depending on model architecture (see \citealp{tesslergoodman2019, yoonetal2020} for some examples).
Parameter estimation is particularly interesting when targeting model parameters that are themselves of some theoretical relevance.
For example, \citet{yoonetal2016} use empirical data to infer plausible values of the parameter $\textbf{x}$ from Equation~\eqref{S1-polite}, which models the relative importance of a speaker's informational and social goals during language production.
\citet{FrankeScholler2016:Semantic-values} infer plausible semantic threshold variables for the vague quantifiers \emph{many} and \emph{few}, one pair for each of several different contexts, in order to address the question of whether it is plausible that the same threshold parameter explains the use of these expressions across different contexts of use.
\citet{schuster2020know} perform a similar Bayesian data analysis to infer semantic variables over uncertainty expressions (e.g., \emph{probably}, \emph{might}), finding that listeners entertain quite a bit of uncertainty about the semantic threshold for uncertainty expressions. \citet{degenetal2020} infer semantic ``noise'' values for color and size modifiers, tracing speakers' propensity to over-modify with color adjectives back to differential semantic noise parameters for size~vs.~color adjectives. 

In addition to estimating model parameters, Bayesian methods can be used to arbitrate between competing hypotheses (so-called ``model comparison''). At a high level, the approach is the same: we are asking which of two (or, $n$) models is the most likely explanation of the observed data. In practice, this aim is accomplished using what are called \emph{Bayes Factors} (or BF for short), which compare the average (marginal) likelihood of the observed data under each model \citep{Jeffreys1961:Theory-of-Proba,KassRaftery1995:Bayes-Factors}. Here, a ``model'' is considered both the RSA model and the prior distribution over parameter values (e.g., that $\alpha$ is sampled from a Uniform prior between 0 and 20). Since BFs take into account the prior distribution over parameters, models with more parameters will have broader prior distributions, and thus the average likelihood of the observed data under such models will intuitively be  watered down by these broad priors (e.g., when the model makes very good predictions but only for a small set of values of the parameters). In this way, BFs take into account model complexity (i.e., more complex models will be penalized if the gains in model fit do not compensate for the increased complexity of the model) when deciding which is the best model.
Examples of comparisons of RSA models based on empirical data include the work of \citet{PottsLassiter2016:Embedded-implic} (albeit in this case the authors do not use Bayesian model comparison), \citet{qingfranke2015}, \citet{FrankeBergen2020:Theory-driven-s}, \citet{yoonetal2020}, and \cite{bohn2019predicting}.

\section{Extensions/limitations} \label{limitations}

With a better sense of the many aspects of probabilistic language understanding that the RSA framework has been used to capture, we turn now to current limitations of the framework and opportunities for further extensions.

\subsection{Compositionality}

All of the models we have considered operate at the utterance level, taking as their starting point whatever the compositional semantics delivers to them as the meaning of a proposition; the models deliberately avoid the composition of the literal interpretations over which they operate. This move is made primarily for convenience. However, one of the most remarkable aspects of natural language is its compositionality: speakers generate arbitrarily-complex meanings by stitching together their smaller, meaning-bearing parts. The compositional nature of language has served as the bedrock of semantic (indeed, linguistic) theory since its modern inception; \cite{montague1973} builds this principle into the bones of his semantics, demonstrating with his fragment how meaning gets constructed from a lexicon and some rules of composition. Since then, compositionality has continued to guide semantic inquiry: what are the meanings of the parts, and what is the nature of the mechanism that composes them? Put differently, what are the representations of the language we use, and what is the nature of the computational system that manipulates them?

We have seen how formalizing pragmatic reasoning in the form of computational cognitive models both informs pragmatic phenomena and enriches theories of semantics. Still, by operating primarily at the level of propositions, the RSA approach necessarily eschews much of the compositional machinery that generates those propositions in the first place. In principle, this semantic leveling is unnecessary; our models of meaning can take into account the rich compositionality of the communicative system they are meant to characterize. The many sources of uncertainty in semantic composition are ripe for a probabilistic treatment, and we now have the tools to deliver one. The basic functionality we are after is already built into the system: lambda abstraction, functional application, declarative memory and thus the possibility for a lexicon.

Indeed, some initial steps have already been taken toward incorporating compositionality within RSA. \cite{goodmanlassiter2015handbook} show how to use the tools of probabilistic programming---the backbone of many RSA applications---to deliver a stochastic lambda calculus that composes the meanings of utterances out of their component parts. \cite{goodmanstuhlmuller2014} go one step further, developing a semantic parsing system based on combinatory categorial grammar that exists internal to the $L_0$ layer of RSA; the parser constructs literal interpretations and verifying worlds from the semantic atoms of sentences. However, neither of these approaches has been scaled up to the full RSA reasoning chain.

By incorporating semantic composition into the RSA framework (rather than approximating it), we further shrink the theoretical and practical distance between semantics and pragmatics by incorporating both within a single model of meaning in language.
Explicitly modeling semantic composition may also help us address the largely ad hoc nature of specifying utterance alternatives inherent to the design of current models. By incorporating a generative grammar, the alternative utterances are those that the grammar may generate; and by dynamically constructing the verifying worlds along the way, we may be able to move past hand-coded priors and toward a more general theory of alternatives and state priors. 
To that end, future work in RSA should examine the ways that a semantic compositional mechanism may be modeled dynamically and probabilistically, within the broader framework of computational cognitive science. A comprehensive approach to modeling language meaning, which treats semantics and pragmatics jointly as a process of probabilistic inference, not only increases the validity of our semantic and pragmatic theories, but more directly informs the psychological underpinnings of language.

\subsection{Adaptation}

In a Bayesian model, the computation of a posterior distribution can be viewed as a specification of the prior distribution in anticipation of the next update.
In other words, ``today's posterior is tomorrow's prior''.
The fact that RSA is a Bayesian model opens the door to formal treatments of adaptation and learning. 

A form of adaptation is at play in the formation of conventions. 
Speakers and listeners must coordinate on how to refer to things in the world.
Such convention formation is studied using so-called ``tangram'' experiments, where participant dyads take turns referring to abstract shapes (the ``tangrams'') with the intention of getting their partner to pick out the same referent. 
Over repeated rounds of this game, participants converge on conventional names for the referents, which are often reduced forms from how they originally refered to them \citep{clark1986referring, hawkins2020characterizing}.
This  lexical adaptation process can be formalized in the RSA framework through interlocutors' reasoning about each others' lexica, as in the meaning-inference models surveyed in Section \ref{meaning-inference} above \citep{hawkins2017convention}.
Beyond the lexicon, adaptation can occur at the level of pragmatics: listeners may adapt to how others are using language (e.g., what utterances certain speakers prefer). 
\cite{schuster2020know} examined listeners' interpretations of uncertainty expressions (\emph{might}, \emph{probably}) and found that listeners update their expectations about the usage of such expressions after minimal evidence from a speaker. 
Using a model-comparison approach within the RSA framework, the authors found that this kind of semantic/pragmatic adaptation is explained by listeners adapting their representations of the speaker's lexicon (i.e., word--meaning mappings) and the utterances a speaker prefers to say. 

\subsection{Language development}

The RSA modeling framework puts forth a particular hypothesis about how language understanding unfolds, and the model components and parameters can be taken to correspond to a hypothesis about the representations that may vary across the population of speakers or develop with age. 
For example, at a young age, children are sensitive to various information sources that help them learn the meanings of new words (e.g., an understanding of common ground, informativity, and their extant lexical knowledge; \citealp{markman1988children, schulze2013, akhtar1996role}).
In naturalistic contexts, multiple information sources may be present, raising the question of how these sources are integrated.
RSA provides a hypothesis about this integration: informativity is captured by the speaker rationality parameter; common ground influences the prior over states.
Building on a paradigm pioneered by \cite{frank2014inferring},
\cite{bohn2019integrating, bohn2019predicting} interrogated this hypothesis by studying 2-to-5 year old children in word learning tasks. 
They use the RSA architecture to formalize a space of hypotheses about the integration process and its development (e.g., a rational integration process, or a biased integration process that deviates from RSA's Bayesian logic).
They find the developmental trajectories in word learning inferences trace back to the improved reliability of the various information sources and not the integration mechanism itself.
Similarly, \cite{savinellietal2017,savinellietal2018} use RSA to model child and adult behavior on ambiguity resolution, thereby providing computational evidence in favor of developmental continuity.

Probabilistic speaker and listener behavior, as captured by RSA models, also helps build more realistic scenarios for language change or language evolution.
Abstractly put, language learners mainly observe pragmatic language use, but need to reconstruct the underlying semantic structure that likely generated the speech they observed.
When this learning or inference task is carried out repeatedly across (stylized) generations, transmission inefficiencies (e.g., through learning biases) can accumulate and shape the cultural evolution of language \citep[e.g.][]{BrochhagenFranke2017:Co-evolution-of,CarcassiSchwoustra2019:The-evolution-o,Carcassi2020:The-cultural-ev}.
\citet{WoensdregtCummins2020:A-computational} apply the combination of iterated learning and RSA modeling to address questions after the potential co-evolution of language and mindreading through numerical simulations.
\citet{OhmerKonig2020:Reinforcement-o} show how pragmatic language use, modelled in terms of RSA speaker and listener protocols, shapes and speeds up the emergence of conventional semantic meaning through reinforcement learning.
\cite{lundetal2019} use RSA to model the engine of pragmatic reasoning in the diachronic development of linguistic meaning.

\subsection{Levels of analysis}

In our characterization of RSA reasoning, we have described agents performing various reasoning steps, talking about $L_1$ reasoning about $S_1$'s reasoning about $L_0$, etc. However, this language can sometimes prove misleading. RSA models are intended to deliver a computational-level description of the problem speakers face as they use language to communicate. As a computational-level analysis \citep{marr1982}, these models intentionally avoid pronouncements about the specific mechanisms involved in solving the problem of language understanding. In other words, the RSA framework is not conceived as a model of language processing.

Indeed, one may wonder about the psychological validity of the rational agents operating within an RSA model.
Language users face limits on their cognitive abilities---limitations of memory, attention, and general processing resources---yet it may not be obvious where such resource limitations enter into RSA.
One avenue through which resource limitations may enter is the optimality parameter $\alpha$, which control's $S_1$'s utility-maximizing behavior.
For example, one may posit that $\alpha$ actually varies across demographic features of the data set (e.g, age of participant); then, one may model $\alpha$ via linear regression (predicting $\alpha$ as a function of age) to see how pragmatic competence develops \citep{bohn2019predicting}.
As $\alpha$ decreases, $S_1$ is less likely to choose actions (i.e., utterances) with higher utility.
Another avenue through which resource limitations may enter RSA is the number of layers of recursive reasoning \citep{FrankeDegen2015:Reasoning-in-Re}.
Moreover, as detailed in Section \ref{prob-programs}, by taking direct inspiration from probabilistic programs, we may see sampling approximations to inference processes as another factor in which RSA-like behavior is a form of resource-bounded rationality \citep{GriffithsLieder2015:Rational-Use-of}.

As a computational-level model operating over full utterances, RSA, at least in the formulations we have considered, also eschews the highly incremental nature of human language processing. Rather than waiting for utterances to terminate before reasoning about them, language users begin integrating information as soon as it arrives during the linear unfolding of time (e.g., \citealp{tanenhausetal1995}). This incremental processing is not limited to word recognition or syntactic parsing; pragmatic reasoning has also been shown to be incrementally engaged (e.g., \citealp{sedivyetal1999,sedivy2007}). We might wonder, then, whether RSA should incorporate incremental pragmatic reasoning, and, if so, how such reasoning may be implemented. \cite{cohngordgonetal2019} offer one path forward for incremental RSA, amending the speaker and listener models so that they reason about individual words and potential continuations, showing how this advance is beneficial for natural language processing applications.
In psycholinguistics, \cite{WerningCosentino2017:The-interaction} and \cite{AugurzkyFranke2019:Gricean-expecta} show how probabilistic predictions about the whole utterance derived from RSA-style modeling can be broken down to next-word expectation, a finding which is supported even by quantitative aspects of EEG measurements.

\subsection{Natural language processing}

The final extension we discuss concerns the utility of RSA reasoning to current applications in language technologies. 
Probabilistic formalizations of Gricean pragmatics lend themselves directly to integration into probabilistic models used in natural language processing.
One line of research focuses on how a semantic meaning representation can be learned from data, based on the assumption that the data was generated by an RSA-style pragmatic agent.
For example, \citet{Monroe:Potts:2015} explore how to set up a model in which RSA-style literal and pragmatic language agents learn underlying semantic meaning associations based on data from the TUNA corpus \citep{DeemterSluis2006:Building-a-Sema}, which provides human data from tasks similar to the reference game from Section~\ref{overview}.
\citeauthor{Monroe:Potts:2015} show that pragmatic agents outperform literal language users in their set-up.
\citet{Monroe:Hawkins:Goodman:Potts:2017} apply a similar strategy of learning semantic representations through the lens of a pragmatic agent to a novel corpus of color descriptions.

Another line of recent research that brings together probabilistic pragmatic reasoning and computational linguistics centers around the realization that we can increase system performance by taking into account a pragmatic perspective for generation and interpretation.
Early work in this vein comes from \citet{AndreasKlein2016:Reasoning-about}, who show how to wrap RSA-style probabilistic reasoning around ``ground-truth'' listener and speaker models, which are trained on human data, and how the resulting pragmatic models improve descriptions of visual scenes in a target-distractor setting.
This general idea of grounding semantic meaning in a data-trained module and building a context-aware pragmatic RSA agent on top of it has been shown to boost performance also in a number of further applications, such as route descriptions \citep{FriedHu2018:Speaker-Followe}, summarization and abstraction \citep{ShenFried2019:Pragmatically-I}, image captioning \citep{Cohn-GordonGoodman2018:Pragmatically-I,NieCohn-Gordon2020:Pragmatic-Issue}, and machine translation \citep{Cohn-GordonGoodman2019:Lost-in-Machine}.

\section{Conclusion} \label{summary}

As the title of the current work promises, our aim has been a practical introduction to the RSA modeling framework: what would the would-be modeler need to design their own computational cognitive model of probabilistic language understanding? We began with a high-level overview of the modeling framework, which treats language understanding as a process of probabilistic, recursive social reasoning; we also stressed how the formalization offers an articulated implementation of Gricean pragmatics. With an understanding of Vanilla RSA from \cite{frankgoodman2012}, which was used to model communication behavior in simple coordinated reference games, we then expanded our sights to a broader range of language phenomena, taking note of the technology used to capture them. We showed how expanding the empirical coverage of RSA often involves taking into account additional sources of uncertainty: uncertainty about the semantics or lexicon, uncertainty about the QUD, uncertainty about the appropriate context, uncertainty about the epistemic states of conversational participants, and uncertainty about the speaker's utility calculus.

After considering the practical and theoretical benefit of thinking through models both ``in math'' and ``in code'', 
we then shifted our focus to issues that commonly arise in model development: how to set parameters and how to link model predictions to the empirical data they are meant to model. While these considerations occupy a significant portion of the modeler's time, they rarely feature prominently in model write-ups, which means that best practices can be hard to identify without an exercise in trial-and-error.

Finally, we explored current limitations of the modeling framework, together with areas where the framework is being actively extended.

Although our focus has been on the modeler, we hope to have offered valuable insight into the framework, which ought to provide even passive consumers with the ability to better evaluate and appreciate the growing RSA literature.

\bibliography{problang}

%\begin{addresses} %%% uncomment to de-anonymize
%	\begin{address}
%		author1
%		\email{email1}
%	\end{address}
%	\begin{address}
%		author2
%		\email{email2}
%	\end{address}
%	\begin{address}
%		author3
%		\email{email3}
%	\end{address}
%\end{addresses}

\end{document}